\documentclass[11pt,a4paper]{article}
\usepackage[hyperref]{acl2021}
\usepackage{times}
\usepackage{latexsym}
\usepackage{graphicx} 
\usepackage{float}  
\usepackage{booktabs} 
\usepackage{multirow} 
\usepackage{multicol}

\usepackage{microtype}

\aclfinalcopy 

\title{Improving Zero-Shot Entity Retrieval through Effective Dense Representations}

\author{
  Eleni Partalidou \\
  Department of Informatics \\
  Aristotle University of Thessaloniki \\
  54124 Thessaloniki – Greece \\
  \texttt{epartalid@csd.auth.gr} \\\And
  Despina Christou \\
  Department of Informatics \\
  Aristotle University of Thessaloniki \\
  54124 Thessaloniki – Greece \\
  \texttt{christoud@csd.auth.gr} 
  \AND
  Grigorios Tsoumakas \\
  Department of Informatics \\
  Aristotle University of Thessaloniki \\
  54124 Thessaloniki – Greece \\
  \texttt{greg@csd.auth.gr} \\}

\date{January 2021}

\begin{document}
\maketitle

\begin{abstract}
Entity Linking (EL) seeks to align entity mentions in text to entries in a knowledge-base and is usually comprised of two phases: candidate generation and candidate ranking. While most methods focus on the latter, it is the candidate generation phase that sets an upper bound to both time and accuracy performance of the overall EL system. 
This work’s contribution is a significant improvement in candidate generation which thus raises the performance threshold for EL, by generating candidates that include the gold entity in the least candidate set (top-K).
We propose a simple approach that efficiently embeds mention-entity pairs in dense space through a BERT-based bi-encoder. Specifically, we extend \cite{wu2020scalable} by introducing a new pooling function and incorporating entity type side-information. We achieve a new state-of-the-art 84.28\% accuracy on top-50 candidates on the Zeshel dataset, compared to the previous 82.06\% on the top-64 of \cite{wu2020scalable}.
We report the results from extensive experimentation using our proposed model on both seen and unseen entity datasets. 
Our results suggest that our method could be a useful complement to existing EL approaches. 

\end{abstract}

\section{Introduction}
\label{intro}

Entity Linking (EL) aims to disambiguate entity mentions in a document against entries in a knowledge base (KB) or a dictionary of entities. Accurately linking these entities plays a key role in various natural language processing (NLP) tasks, including information extraction \cite{lin2012entity, hasibi2016exploiting}, KB population \cite{dredze2010entity}, content analysis \cite{huang2018entity} and question answering \cite{li2020efficient}, with potential applications in many fields, including technical writing, digital humanities, and biomedical data analysis.

While EL systems typically rely on external KBs and assume entities at inference time known, real-world applications are usually accompanied by minimal to zero labeled data, highlighting the importance of approaches that can generalize to unseen entities.  \citet{logeswaran2019zero} introduced Zero-shot EL, where mentions must be linked to unseen entities, without in-domain labeled data, given only the entities' text description. By comparing two texts - a mention in context and a candidate entity description - the task goes closer to reading comprehension.

Most EL systems consist of two subsystems: Candidate Generation (CG), where for each entity mention the system detects entities related to the mention and document, and Candidate Ranking (CR) where the system chooses the most probable entity link among the found candidates.
Most state-of-the-art models \cite{logeswaran2019zero, li2020efficient, yao2020zero} rely on traditional frequency-based CG and focus on building robust candidate rankers using cross-encoders to jointly encode mention and entity candidate descriptions. However, the memory-intensive CR phase depends on the set of candidates provided by CG. Therefore, the accuracy of CR (and, by extension, the overall EL system), is capped by that of the CG phase.

The goal of this work is to advance the state of the art in the CG phase \cite{wu2020scalable}, in order to set a higher performance {\em ceiling} for CR and for EL overall. We start from \cite{wu2020scalable}, which encodes mention and entity descriptions in dense space through a BERT-based bi-encoder and then retrieves the closest $K$ entities to each mention using the dot product method. Although this method increased accuracy at top-64 retrieved candidates from 69.13\% \cite{logeswaran2019zero} to 82.06\%, an approach that allows higher CG accuracy with fewer candidates is yet to be developed.

We introduce a new state-of-the-art CG model that achieves 84.28\% accuracy on the top-50 candidates; namely, we obtain a 2.22 higher unit threshold in CR and EL overall using 21.88\% fewer candidates, compared to \cite{wu2020scalable}.
Our proposed model uses a BERT-based bi-encoder and incorporates a special pooling function that efficiently encodes mention and entity descriptions in dense space by aggregating information from special tokens. 
We experimented on various pooling function alternatives, retrieval methods, and the effect of entity type side-information. At the same time, we report our results on both seen and unseen entity sets to evaluate our model's efficiency on both typical and zero-shot sets of EL. 
Our results revealed that concatenating special tokens of the proposed BERT-based input representation and using dot product to retrieve candidates yield the best outcomes on both \textit{test} and \textit{heldout\_train\_seen} sets. Meanwhile, state-of-the-art results were preserved without the entity type information, warranting our model's robustness regardless of additional information.

The contributions of this paper are summarized as follows:
\begin{itemize}
    \item We introduce a new CG approach, based on BERT-based bi-encoder and a new pooling function that manages to encode mention and entities in the same dense space very effectively, achieving state-of-the-art results.
    \item A 84.28\% accuracy on top-50 candidates, compared to 82.06\% at top-64 of \cite{wu2020scalable}, measured on the Zeshel test dataset. We thus increase the CR accuracy by 2.22 units while requiring 21.88\% fewer candidates, allowing more accurate and faster inference.
    \item State-of-the-art results on both seen and unseen entity sets, with and without entity type information, reveal the robustness of our model in both the typical and the zero-shot EL settings.
\end{itemize}


 


\section{Related Work}
\label{related}
Former work has pointed out the importance of building entity linking systems that can generalize to unknown named entities, either via a reduced candidate set or through a more robust candidate ranker. Our work 
falls into the first category, namely the candidate generation phase and is related to BERT-based sequence modeling, pooling techniques, entity type side-information, and candidate retrieval methods.


\paragraph{Candidate Generation (CG)}
Traditional CG has been based on string comparison \cite{phan2017neupl} and alias tables. These methods lack rich representation and restrict to a small entity set, respectively, thus proving inefficient for the challenging task of entity candidates retrieval, especially in the zero-shot set, where generalization to new entities is on the spot.
Over the last years, many works \cite{sil2012linking, murty2018hierarchical, logeswaran2019zero} focused on frequency-based methods, with most of them following TF-IDF and BM25 approaches. \citet{gillick2019learning} introduces a simple neural bi-encoder and shows that encoding mention and entities in dense space works well. Inspired by this idea, \cite{wu2020scalable} proposed the, to our knowledge, current state-of-the-art CG model, using a more robust transformer-based bi-encoder \cite{humeau2019poly}. Precisely, they achieve an 82.06\% acc. on top-64 retrieved candidates in the Zeshel dataset. Their model uses a BERT-based bi-encoder to encode mention and entities' descriptions in dense space, where the top-K entities are then retrieved based on the two vectors' maximum dot product.
Our work extends the work of \cite{wu2020scalable} by investigating other than the default BERT \cite{devlin2019bert} pooling functions, additional entity type side-information, and alternate retrieval methods.

\paragraph{Entity Type side-information}
Apart from model architecture, several methods \cite{raiman2018deeptype, khalife2018scalable} show that fine-grained entity type side-information helps to further disambiguate entities in EL. First attempts to integrate type information in the task consisted of end-to-end systems \cite{durrett2014joint, luo2015joint, nguyen2016j} that jointly modeled entity linking and named entity recognition to capture the mutual dependency between them, typically using structured CRF and hand-crafted features. Automatically extracted entity features were captured by \cite{yamada2016joint,ganea2017deep} using typical embedding methods proposed by \cite{mikolov2013distributed}, while \cite{martins2019joint} learned entity features using Stack-LSTMs \cite{dyer2015transition} in multi-task learning. Latest works in EL \cite{hou2020improving, chen2020improving} suggest to build entity type embeddings by modeling the mentions' near context and its semantics using Word2Vec \cite{mikolov2013distributed} and BERT \cite{devlin2019bert} embeddings, respectively.
In our work, we choose to model entities as special tokens in BERT representation and let the model automatically capture relevant information across all sequence, following similar BERT-based entity representations \cite{shi2019simple, poerner2020bert, christou2021improving} applied in other NLP applications.

\paragraph{Candidate Retrieval methods}
As mentioned above, most approaches \cite{logeswaran2019zero, sil2012linking} retrieve entity candidates using frequency-based methods, such as BM25 \cite{robertson2009probabilistic}. Meanwhile, recent bi-encoder approaches in EL \cite{gillick2019learning, wu2020scalable} score mention-candidate pairs using dot product during training while retrieving top-K candidates using the same vector similarity measure. In our work, we experiment on alternate candidate retrieval methods, i.e., euclidean, cosine, and dot methods, to find out cosine \cite{manning2008prabhakar} and dot product to outperform the euclidean metric significantly.


\section{Methodology}
Our proposed method for candidate generation uses BERT \cite{devlin2019bert} in a bi-encoder set. The bi-encoder uses two independent BERT transformers to encode the mention's context and all entity descriptions into two dense vectors. Then, the best candidates are retrieved by scoring the mention-entity pair vectors with a vector similarity measure. This section presents specific sequence representation methods, and other than the typical dot-product metric used to retrieve the best candidates.

\subsection{Bi-encoder}
\label{bi_encoder_representation}
We use a BERT-based bi-encoder, following \cite{wu2020scalable} to model the mention-entity pairs, because it allows for fast, real-time inference, as the candidate representations can be cached. Below, we present our structured input to the bi-encoder, our proposed pooling functions for sequence representation and brief notes on entity type incorporation in the structured input.

\paragraph{Input Representation}
The mention context and the entity description are encoded into the following vectors:

$$ {\bf y_m} = red(T_1(t_m)) $$ $${\bf y_e} = red(T_2(t_e)) $$

where $t_m$ is the representation of mention, $t_e$ is the representation of entity, $T_1$ and $T_2$ are two transformers. The function red(.) reduces the sequence of vectors into one. By default, the last layer of the output of the [CLS] token is returned.

The representation $t_m$ is composed of the mention text surrounded by two special tokens that denote the existence of the mention, the context from the left and the right side of the mention text plus the special tokens that BERT inserts. Specifically, the construction of the mention is:

\[ [CLS]\;ctxtl\;[Ms]\;mention\;[Me]\;ctxtr\;[SEP] \]

and with the addition of the entity type it is:

\[ [CLS]\;[ent\_type]\;mention [H\_SEP]\;ctxtl ... \]

where [Ms] and [Me] are the special tokens that tag the mention, ctxtl and ctxtr are the word-pieces tokens of the context before and after the mention and [CLS] and [SEP] are BERT's special tokens.

The representation $t_e$ is composed of the entity title, the special token [ENT] separating the entity's title and description plus the special tokens that BERT inserts.

\[ [CLS]\;title\;[ENT]\;description\;[SEP] \]

and with the addition of the entity type it is:

\[ [CLS]\;[ent\_type]\;title ... \]

For simplicity in both input representations a maximum length of the document is retrieved.

The score of the entity candidate $e_i$ given a mention is computed by the dot-product:

$$ s(m,e_i) = {\bf y_m} {\cdot} {\bf y_{ei}} $$

The network is trained to maximize the score of the correct entity with respect to the entities of the same batch. For  each  training  pair $(m_i,e_i)$ in  a batch of B pairs, the loss is computed as:

$$ L(m_i, e_i) = -s(m_i, e_i) + log(\sum_{j=1}^{B} exp(s(m_i, e_j))) $$







\paragraph{Input Encoding}
The dataset needed to be processed by the models to understand information and to extract representative vectors from it. For that reason, a Tokenizer was used to split each document into a list of tokens and to give each token a respective id.
The Bert Tokenizer used a special technique called Byte Pair Encoding \cite{sennrich2016neural}. BPE makes use of a simple data compression technique that iteratively replaces the most frequent pair of bytes in a sequence with a single, unused byte. Instead of merging frequent pairs of bytes, characters or character sequences are merged. The reason this technique is used by the algorithm is to compute subwords from unknown tokens, so words that the algorithm may known can be associated with unknown words that tend to be similar to them and consequently are given proper token ids.

\paragraph{Pooling Functions}
The different sequence representations that were tested fall into the following categories:
\begin{itemize}
    \item Usage of [CLS]: The default representation of the sequence.  $$ CLS = {h_L}[0] $$
    
    \item Average of tokens: Averages all token representations from BERT's last hidden layer. 
    $$ avg = \frac{\sum_{i=0}^{n}(h_L[i])}{n} $$
    
    \item Sum of tokens: Sums all token representations from BERT's last hidden layer. 
    $$ sum = \sum_{i=0}^{n}(h_L[i]) $$
    
    \item Average\, special: Averages only the special token representations from BERT's last hidden layer.  
    $$ avg\,special = \frac{\sum_{i=0}^{n}(h_L[i])}{n} $$
    
    \item Sum of special tokens: Sums only the special token representations from BERT's last hidden layer.   
    $$ sum\,special =  \sum_{i=0}^{n}(h_L[i]) $$
    
    \item Concatenation of special tokens: Concatenates only the special token representations from BERT's last hidden layer.  
    $$ conc\,special = [h_L[j_1], ..., h_L[j_M]],$$ where $j$ refers to the index of special tokens with $M$ being the total number of special tokens.
\end{itemize}

We note that $h_L$ corresponds to the last hidden layer of BERT, $n=128$, namely the max length, and special tokens declare specific parts of the sequence (i.e. $[CLS]$, $[SEP]$, $[M_s]$, $[M_e]$, $[ent\_type]$) that are not split by the tokenizer. 

\paragraph{Entity Type Representation}
In the extent that entity types put some  constraint on the possible entity to link, we also experimented on incorporating the entity type in the model's structured input. 
Precisely, we incorporate 18 generic entity types, captured from recognizing mention's and entity titles' entity types with the spaCy model\footnote{\url{https://github.com/explosion/spacy-models/releases/tag/en_core_web_lg-2.3.0}}. SpaCy is an open-source software library for advanced natural language processing, written in the programming languages Python and Cython \cite{spacy}.
We note that entity types, include the $<unk>$ type for unclassified mentions/entity titles. Also, all entity types are encoded as special tokens; consequently, sequence representations that make use of special tokens also consider the entity type embeddings.

\subsection{Retrieval Methods}
In the candidate generation step different methods were performed, in order to find a proper candidate set. The first method was to initialize a k-nearest neighbors classifier and fit the model with the representations of the entity descriptions from the bi-encoder and their respective id. After that, for each context mention the k closest neighbors were returned, where k the number of candidates the candidate set should have.

Another approach was to compute the similarity between each context mention with each entity and construct the candidate set with the h more common ones. To find the similarity between two vectors ${\bf A} = [a_1,a_2,...,a_n]$ and ${\bf B} = [b_1,b_2,...,b_n]$ , some main similarity measures were used to choose from.

The cosine similarity method measures the cosine of angle $\theta$ between two vectors. The cosine increases as the two vectors tend to look alike.

$$ \cos ({\bf A},{\bf B}) = \frac{{\bf A}^T {\bf B}}{ |{\bf A}| {\cdot |{\bf B}|}}  $$

The dot product method measures the cosine multiplied by lengths of both vectors. For very similar vectors the value is increased and also for vectors with big length.

$$ {\bf A} \cdot {\bf B} =  a_1b_1+a_2b_2+...+a_nb_n =|{\bf A}||{\bf B}|cos(\theta) $$

The euclidean distance method measures the distance between ends of vectors. For vectors that are similar and consequently very close to the vector space the value is small.
 
$$ d({\bf A}, {\bf B}) = \sqrt{(a_1-b_1)^2+...+(a_N-b_N)^2} $$

\section{Experimental Setup}
\label{experimental_setup}

\subsection{Dataset}
For our experiments, we use the Zeshel\footnote{https://github.com/lajanugen/zeshel} dataset, which to our knowledge, is the prevailing benchmark for zero-shot EL. The dataset was constructed using documents from Wikias, namely community-written encyclopedias, each specialized in a particular subject. Documents in Wikias exhibit rich document context, thus are an excellent source for natural language comprehension models. Meanwhile, each encyclopedia contains domain-specific entity dictionaries, making it an ideal dataset for evaluating the EL system's generalization on new domains.  

Table \ref{zeshel_dataset_stats} presents the number of unique entities per domain (world) and {train/val/test} sets. Regarding mentions, the training set includes 49,275 labeled mentions, while the validation and test sets consist of 10,000 and 10,000 unseen mentions, respectively. Moreover, an extra 5,000 seen and 5,000 unseen mentions from the training set are provided. The goal of this discrimination is to explore the models' performance to known domains. We present our model's performance on both known and unknown worlds in section \ref{case_study}.

\begin{table}[ht!]
\centering
\begin{tabular}{lll}
\textbf{Set} & \textbf{World} & \textbf{Entities} \\
\hline
Train & American Football & 31929 \\
 & Doctor Who &  40281 \\
 & Fallout & 16992 \\
 & Final Fantasy & 14044 \\ 
 & Military & 104520 \\
 & Pro Wrestling & 10133 \\
 & StarWars & 87056 \\
 & World of Warcraft & 27677 \\
\hline
Val & Coronation Street & 17809 \\
 & Muppets & 21344 \\
 & Ice Hockey & 28684 \\
 & Elder Scrolls & 21712 \\
\hline
Test & Forgotten Realms & 15603 \\
 & Lego & 10076 \\
 & Star Trek & 34430 \\
 & YuGiOh & 10031 \\
\hline
\end{tabular}
\caption{Number of Entities per world and {train, val, test} sets in the Zeshel dataset.}
\label{zeshel_dataset_stats}
\end{table}


\subsection{Hyper-parameter Settings}

In our experiments we utilize \textit{bert-base-uncased} model with hidden layer dimension $D_h=768$, while we fine-tune the model with \textit{max\_seq\_length} $D_t=128$. Regarding model's hyper-parameters, we manually tune them on the training set. For fine-tuning, we assigned $batch\_size=8$, $epochs=5$, all the BERT's layers are updated during back-propagation, learning rate $lr=3e^{-5}$ and $weight\_decay=0.01$. Moreover, we fine-tune our model using the Adam optimization scheme \cite{loshchilov2017decoupled} with $\beta_1 = 0.9, \beta_2 = 0.999$ and a linear learning rate decay schedule. We minimize loss using cross entropy criterion.

Experiments were conducted on a PC with 15 GB RAM, an AMD FX-8350 Eight-Core Processor with a base frequency of 4.00 GHz and a NVIDIA GeForce GTX TITAN X graphics card with 12 GB memory.

Training time took about 150min for the whole process. For the CG task and to process all test set (10K mentions, 70.140 entities), the model takes 90min using cosine method, 60min with the euclidean distance method and 50min with the dot product method. For the heldout\_train\_seen set, the model takes 30min with the cosine method, 25min with the euclidean distance method and 20min with the dot product method.

\subsection{Evaluation}
To evaluate the performance of our model against the rest state-of-the-art models, we report the accuracy at top-k. Namely, we assess the performance on the subset of test instances for which the gold entity is among the top-k retrieved candidates. We highlight that since EL systems' overall performance is upper-bounded by the CG accuracy and their speed is associated with the number of retrieved candidates, a model that will increase CG performance in fewer candidate entities is crucial.


\subsection{State-of-the-art CG Models}
For evaluating our method, we compare against the following state-of-the-art models in Zero-Shot EL:
\newline
\textbf{BM25} \cite{logeswaran2019zero}: Use a traditional, frequency-based method (BM25) for the candidate generation. To re-rank the top K candidates propose the use of a cross-encoder to jointly encode mention context and entity description.
\newline
\textbf{Bi-encoder} \cite{wu2020scalable}: First to perform the candidate retrieval on a dense space. Mention context and candidate entities are encoded into vectors using a bi-encoder and the retrieval phase is then reduced to finding the maximum dot product between mention and entity candidate representations. Final entity selection is performed using the top K candidates from the retrieval phase and re-ranked through a cross-encoder, similar to \cite{logeswaran2019zero}. As an encoder BERT model is selected.
\newline
While state-of-the-art models in Zero-shot EL \cite{logeswaran2019zero, yao2020zero} focus on the CR phase, \citet{wu2020scalable} are the only to propose a different to traditional IR approach for CG. Our focus is to further push the boundaries of the CG phase and set a higher performance threshold to CR and EL overall.

\section{Results}
\label{results}

\subsection{Comparison with Baselines}

\begin{table}[h!]
\centering
\begin{tabular}{lcc}
\hline
\textbf{Method} & \textbf{Test} & \textbf{Top-K}  \\
\hline
\cite{logeswaran2019zero}  & 69.13 & 64  \\
\cite{wu2020scalable}      & 82.06 & 64  \\
\hline
Ours (sum)$\dagger$         & 80.27 & 50   \\
Ours (sum special)$\dagger$ & 83.23 & 50   \\
Ours (CLS)$\dagger$         & 83.83 & 50   \\
Ours (avg special)$\dagger$ & 83.88 & 50   \\
Ours (avg)$\dagger$         & 84.19 & 50   \\
\hline
Ours (conc special) & \textbf{84.28} & 50   \\
\hline
\end{tabular}
\caption{\label{BaselinesComparison} Normalized accuracy on the test set of Zero-shot EL dataset. Baseline models are compared against our model's various sequence representation alternatives, with selected pooling function noted in parentheses. $\dagger$ indicates methods including entity type side-information.}
\end{table}

Here, we compare our CG model against the previous state-of-the-art works and our model's various alternatives that use different pooling functions for sequence representation and incorporate or not entity type side-information. Table \ref{BaselinesComparison} shows our model to significantly outperform the previous state-of-the-art works with fewer candidates regardless of chosen pooling function and additional side-information. Precisely, the best version of our model achieves 2.22 and 15.15 higher normalized accuracy than \cite{wu2020scalable} and \cite{logeswaran2019zero} respectively, using 21.88\% fewer candidates and without incorporating any additional side-information.  

\begin{figure}[h!]
\centering\includegraphics[scale=0.63]{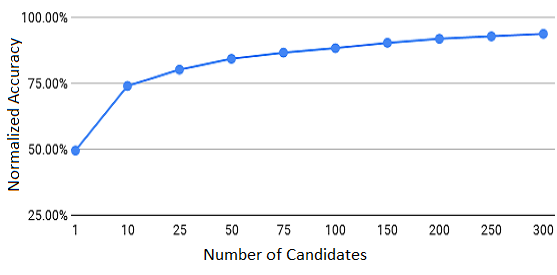}
\caption{Top-K entity retrieval accuracy of \textit{Ours (conc special)} model.}
\label{Topk_CG_accuracy_concat_cpecial_no_entity_type}
\end{figure}

Meanwhile, in Figure \ref{Topk_CG_accuracy_concat_cpecial_no_entity_type}, which shows our best model's accuracy across various top-K retrieved candidates, we can see that choosing more candidates could further increase accuracy. Still, we decided on fifty (50) candidates due to its small enough size and performance improvement over the rest baselines.  


\subsection{Ablation Studies}
In this section, we review the effect of pooling functions, entity type side-information and retrieval methods.

\subsubsection{Effect of Entity Type across all Pooling Functions}
We experimented with various pooling functions to find the most efficient sequence representation, whereas for each representation, we explore the addition of the mention's entity type at the sequence encoding (see section \ref{bi_encoder_representation}).

\begin{table}[H]
\centering
\begin{tabular}{lcc}
\hline
 & \multicolumn{2}{c}{\textbf{Test}} \\
\textbf{Pooling Functions} & w/o Ent.Type & Ent.Type \\
\hline
Sum tokens            & 79.64\% & \textbf{80.27\%} \\
CLS token             & 83.52\% & \textbf{83.83\%} \\
Avg special tokens    & 83.83\% & \textbf{83.88\%} \\
Avg tokens            & 84.06\% & \textbf{84.19\%} \\
Conc special tokens   & \textbf{84.28}\% & 83.30\% \\
\hline
\end{tabular}
\caption{\label{dot_test} Comparison of results with and without entity type side-information for each pooling function on the Test set of Zero-shot EL dataset. We report results on the top-50 candidates, using the dot product as retrieval method.}
\end{table}

Table \ref{dot_test} presents the normalized accuracy on top-50 candidates for all pooling functions presented in section \ref{bi_encoder_representation} incorporating or not the entity type side-information. As can be seen, state-of-the-art accuracy is achieved by representing the sequence as the concatenation of its special tokens without incorporating entity-type information. 
Meanwhile, the addition of entity type in the rest representation techniques seems to improve performance, but only slightly (up to 0.63\%).
We suspect that the reason that entity type does not add significant value is that about 60\% of mentions are detected with UNKNOWN entity type; thus, the model was not able to capture robust patterns. From manual inspection, we notice that most UNKNOWN entity types are associated with pronouns or co-references.

\subsubsection{Effect of the Retrieval Methods}
Regarding the retrieval phase, we wanted to explore the effect of other common distance metrics, apart from the typical one used, i.e., the dot product. 

\begin{table}[h!]
\centering
\begin{tabular}{lccc}
\hline
\textbf{Method/} & \multicolumn{1}{c}{} & {\textbf{Test}} & \multicolumn{1}{c}{}  \\
\textbf{Distance Metrics}  & \multicolumn{1}{c}{Euclid.} & \multicolumn{1}{c}{Cosine} & \multicolumn{1}{c}{Dot}  \\
\hline
Ours (CLS) & 79.51 & 82.74 & 83.52  \\
Ours (avg) & 81.27 & 83.73 &  84.06 \\
Ours (sum) & 76.05 & 78.02 &  79.64 \\
Ours (avg special) &  82.08 & 83.55 & 83.83 \\
Ours (sum special) & 81.56 & 82.36 &  82.90 \\
Ours (conc special) & 79.96 & 83.63 & \textbf{84.28} \\
\hline
\end{tabular}
\caption{\label{CG_results} Comparison of all methods using Euclidean, Cosine and Dot product distance metrics on the top-50 retrieved candidates on the test set of the Zero-shot EL dataset. Reported results refer to Normalized Acc., while all methods do not incorporate entity type.}
\end{table}

\begin{figure}[h!]
\centering\includegraphics[scale=0.32]{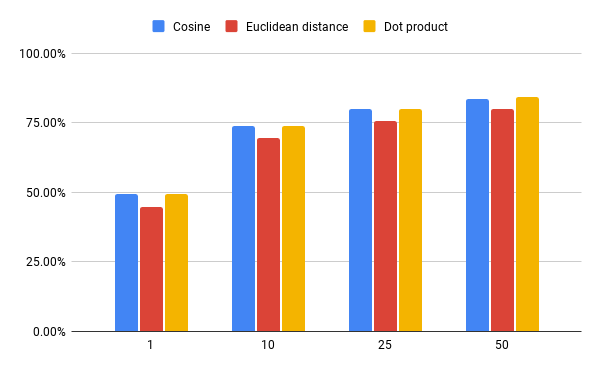}
\caption{Performance of \textit{(conc special)} model for all Euclidean, Cosine, and Dot product distance metrics across the top retrieved candidates.}
\label{concat_special_wo_entity_type_retrieval_methods_evaluation}
\end{figure}

Table \ref{CG_results} compares all methods' performance using the Euclidean, Cosine, and Dot product distance metrics to retrieve the best candidates. The Euclidean distance presents the worst performance for all methods, with dot product consistently slightly outperform the cosine. 
Figure \ref{concat_special_wo_entity_type_retrieval_methods_evaluation}, which shows our best model's performance for all three distance metrics across top-\{1, 10, 25, 50\} retrieved candidates, exhibits similar observations. Namely, the dot product achieves higher accuracy across all top candidates, closely followed by cosine and then by the euclidean distance. 
Consequently, our findings are in sync with \cite{manning2008prabhakar}. Comparing the three distance metrics, cosine and dot product metrics are advantageous to euclidean as they measure the angle between documents; namely, they measure documents' similarity regardless of their size. Moreover, the dot product considers both the angle and the magnitude of two vectors, which makes it competitive to the cosine similarity metric.

\subsection{Case Study: Performance on the typical EL set}
\label{case_study}
So far, we have tested our methods' efficiency in the most challenging set of EL, the zero-shot set, where we must prove the model's generalization in out-of-domain, unseen entities. In this section, our scope is to examine the in-domain generalization performance and prove our method as a valuable complement to all existing EL approaches. 

To test the performance on in-domain, seen entities, we exploit the \textit{``Heldout train seen"} set, provided by \cite{logeswaran2019zero}, that includes 5,000 mentions linked to seen entities during training. In \autoref{dot_product_heldout_train_seen}, we present the normalized accuracy on top-50 candidates for all pooling functions incorporating or not the entity type side-information, whereas in \autoref{similarity_functions_heldout_train_seen} we show the performance of our best model using different distance metrics in the retrieval phase.

\begin{table}[h!]
\centering
\begin{tabular}{lcc}
\hline
 & \multicolumn{2}{c}{\textbf{Heldout Train Seen}} \\
\textbf{Pooling functions} & w/o Ent.Type & Ent.Type \\
\hline
[CLS]                  & 98.42\% & 98.56\% \\
Avg tokens             & 98.56\% & 98.05\% \\
Sum tokens             & 97.60\% & 97.62\% \\
Avg special tokens     & 98.76\% & 98.54\% \\
Sum special tokens     & 98.26\%  & 98.30\% \\
Conc special tokens    & \textbf{98.86\%} & 98.36\% \\
\hline
\end{tabular}
\caption{\label{dot_product_heldout_train_seen} Comparison of results with and without entity type for each pooling function on the \textit{Heldout Train Seen} set. We report results on the top-50 candidates, using the dot product as retrieval method.}
\end{table}

Both tables reveal same observations to our methods' performance in the Zero-shot set. Namely, using the concatenation of special tokens to represent sequences and the dot product to retrieve the best candidates leads to greater accuracy. Moreover, entity type only slightly boosts few sequence representation methods, and euclidean distance performs the poorest. At last, we provide a great upper threshold (98.86\%) for EL under its typical, supervised set.

\begin{table}[H]
\centering
\begin{tabular}{lcc}
\hline
\textbf{Distance  metrics} & \textbf{Results} \\
\hline
Euclidean     & 97.3\% \\
Cosine        & 98.64\% \\
Dot product   & \textbf{98.86\%} \\
\hline
\end{tabular}
\caption{\label{similarity_functions_heldout_train_seen} Comparison of the similarity methods for concatenation of special tokens and 50 candidates. Results for the heldout train seen set.}
\end{table}

\section{Conclusion}


We proposed a simple yet very effective candidate generation model that sets a higher performance bar for candidate ranking (CR) and EL overall. Precisely, our model achieves at least 2.22 higher threshold for CR and EL using 22\% fewer candidates, compared to previous state-of-the-art works in the zero-shot set. Meanwhile, we tested our model's efficiency with in-domain corpus, seen entities to also accomplish great results with the same parameters as in zero-shot EL, proving that the suggested method can be a valuable complement to all existing EL approaches. 
We also experimented with adding entity types to boost performance, without success, though. After inspecting the data, we found out that 60\% of the mentions were not classified with an entity type, with most of them constituting pronouns, not named entities. On the one hand, that proves our model's robustness regardless of side-information, but from the other suggest specific enhancements for the future. Future enhancements could incorporate entity graph information instead of entity types to capture co-references better and integrate a generic document entity type to locate the main concept of the document. At last, we plan to extend this work with respective experimentation in the candidate ranking phase.



\bibliographystyle{acl_natbib}

\begin{thebibliography}{34}
\expandafter\ifx\csname natexlab\endcsname\relax\def\natexlab#1{#1}\fi

\bibitem[{Chen et~al.(2020)Chen, Wang, Jiang, and Lin}]{chen2020improving}
Shuang Chen, Jinpeng Wang, Feng Jiang, and Chin-Yew Lin. 2020.
\newblock Improving entity linking by modeling latent entity type information.
\newblock In \emph{Proceedings of the AAAI Conference on Artificial
  Intelligence}, volume~34, pages 7529--7537.

\bibitem[{Christou and Tsoumakas(2021)}]{christou2021improving}
Despina Christou and Grigorios Tsoumakas. 2021.
\newblock Improving distantly-supervised relation extraction through bert-based
  label \& instance embeddings.
\newblock \emph{arXiv preprint arXiv:2102.01156}.

\bibitem[{Devlin et~al.(2019)Devlin, Chang, Lee, and
  Toutanova}]{devlin2019bert}
Jacob Devlin, Ming-Wei Chang, Kenton Lee, and Kristina Toutanova. 2019.
\newblock Bert: Pre-training of deep bidirectional transformers for language
  understanding.
\newblock In \emph{Proceedings of the 2019 Conference of the North American
  Chapter of the Association for Computational Linguistics: Human Language
  Technologies, Volume 1 (Long and Short Papers)}, pages 4171--4186.

\bibitem[{Dredze et~al.(2010)Dredze, McNamee, Rao, Gerber, and
  Finin}]{dredze2010entity}
Mark Dredze, Paul McNamee, Delip Rao, Adam Gerber, and Tim Finin. 2010.
\newblock Entity disambiguation for knowledge base population.
\newblock In \emph{Proceedings of the 23rd International Conference on
  Computational Linguistics (Coling 2010)}, pages 277--285.

\bibitem[{Durrett and Klein(2014)}]{durrett2014joint}
Greg Durrett and Dan Klein. 2014.
\newblock A joint model for entity analysis: Coreference, typing, and linking.
\newblock \emph{Transactions of the association for computational linguistics},
  2:477--490.

\bibitem[{Dyer et~al.(2015)Dyer, Ballesteros, Ling, Matthews, and
  Smith}]{dyer2015transition}
Chris Dyer, Miguel Ballesteros, Wang Ling, Austin Matthews, and Noah~A Smith.
  2015.
\newblock Transition-based dependency parsing with stack long short-term
  memory.
\newblock In \emph{Proceedings of the 53rd Annual Meeting of the Association
  for Computational Linguistics and the 7th International Joint Conference on
  Natural Language Processing (Volume 1: Long Papers)}, pages 334--343.

\bibitem[{Ganea and Hofmann(2017)}]{ganea2017deep}
Octavian-Eugen Ganea and Thomas Hofmann. 2017.
\newblock Deep joint entity disambiguation with local neural attention.
\newblock In \emph{Proceedings of the 2017 Conference on Empirical Methods in
  Natural Language Processing}, pages 2619--2629.

\bibitem[{Gillick et~al.(2019)Gillick, Kulkarni, Lansing, Presta, Baldridge,
  Ie, and Garcia-Olano}]{gillick2019learning}
Dan Gillick, Sayali Kulkarni, Larry Lansing, Alessandro Presta, Jason
  Baldridge, Eugene Ie, and Diego Garcia-Olano. 2019.
\newblock Learning dense representations for entity retrieval.
\newblock In \emph{Proceedings of the 23rd Conference on Computational Natural
  Language Learning (CoNLL)}, pages 528--537.

\bibitem[{Hasibi et~al.(2016)Hasibi, Balog, and
  Bratsberg}]{hasibi2016exploiting}
Faegheh Hasibi, Krisztian Balog, and Svein~Erik Bratsberg. 2016.
\newblock Exploiting entity linking in queries for entity retrieval.
\newblock In \emph{Proceedings of the 2016 ACM International Conference on the
  Theory of Information Retrieval}, pages 209--218.

\bibitem[{Honnibal et~al.(2020)Honnibal, Montani, Van~Landeghem, and
  Boyd}]{spacy}
Matthew Honnibal, Ines Montani, Sofie Van~Landeghem, and Adriane Boyd. 2020.
\newblock \href {https://doi.org/10.5281/zenodo.1212303} {{spaCy:
  Industrial-strength Natural Language Processing in Python}}.

\bibitem[{Hou et~al.(2020)Hou, Wang, He, and Zhou}]{hou2020improving}
Feng Hou, Ruili Wang, Jun He, and Yi~Zhou. 2020.
\newblock Improving entity linking through semantic reinforced entity
  embeddings.
\newblock In \emph{Proceedings of the 58th Annual Meeting of the Association
  for Computational Linguistics}, pages 6843--6848.

\bibitem[{Huang et~al.(2018)Huang, Cautis, Cheng, Zheng, Mamoulis, and
  Yan}]{huang2018entity}
Zhipeng Huang, Bogdan Cautis, Reynold Cheng, Yudian Zheng, Nikos Mamoulis, and
  Jing Yan. 2018.
\newblock Entity-based query recommendation for long-tail queries.
\newblock \emph{ACM Transactions on Knowledge Discovery from Data (TKDD)},
  12(6):1--24.

\bibitem[{Humeau et~al.(2019)Humeau, Shuster, Lachaux, and
  Weston}]{humeau2019poly}
Samuel Humeau, Kurt Shuster, Marie-Anne Lachaux, and Jason Weston. 2019.
\newblock Poly-encoders: Architectures and pre-training strategies for fast and
  accurate multi-sentence scoring.
\newblock In \emph{International Conference on Learning Representations}.

\bibitem[{Khalife and Vazirgiannis(2018)}]{khalife2018scalable}
Sammy Khalife and Michalis Vazirgiannis. 2018.
\newblock Scalable graph-based individual named entity identification.
\newblock \emph{arXiv preprint arXiv:1811.10547}.

\bibitem[{Li et~al.(2020)Li, Min, Iyer, Mehdad, and Yih}]{li2020efficient}
Belinda~Z Li, Sewon Min, Srinivasan Iyer, Yashar Mehdad, and Wen-tau Yih. 2020.
\newblock Efficient one-pass end-to-end entity linking for questions.
\newblock In \emph{Proceedings of the 2020 Conference on Empirical Methods in
  Natural Language Processing (EMNLP)}, pages 6433--6441.

\bibitem[{Lin et~al.(2012)Lin, Etzioni et~al.}]{lin2012entity}
Thomas Lin, Oren Etzioni, et~al. 2012.
\newblock Entity linking at web scale.
\newblock In \emph{Proceedings of the Joint Workshop on Automatic Knowledge
  Base Construction and Web-scale Knowledge Extraction (AKBC-WEKEX)}, pages
  84--88.

\bibitem[{Logeswaran et~al.(2019)Logeswaran, Chang, Lee, Toutanova, Devlin, and
  Lee}]{logeswaran2019zero}
Lajanugen Logeswaran, Ming-Wei Chang, Kenton Lee, Kristina Toutanova, Jacob
  Devlin, and Honglak Lee. 2019.
\newblock Zero-shot entity linking by reading entity descriptions.
\newblock In \emph{Proceedings of the 57th Annual Meeting of the Association
  for Computational Linguistics}, pages 3449--3460.

\bibitem[{Loshchilov and Hutter(2017)}]{loshchilov2017decoupled}
Ilya Loshchilov and Frank Hutter. 2017.
\newblock Decoupled weight decay regularization.
\newblock \emph{arXiv preprint arXiv:1711.05101}.

\bibitem[{Luo et~al.(2015)Luo, Huang, Lin, and Nie}]{luo2015joint}
Gang Luo, Xiaojiang Huang, Chin-Yew Lin, and Zaiqing Nie. 2015.
\newblock Joint entity recognition and disambiguation.
\newblock In \emph{Proceedings of the 2015 Conference on Empirical Methods in
  Natural Language Processing}, pages 879--888.

\bibitem[{Manning(2008)}]{manning2008prabhakar}
Christopher~D Manning. 2008.
\newblock Prabhakar raghavan, and hinrich schutze.
\newblock \emph{Introduction to information retrieval}.

\bibitem[{Martins et~al.(2019)Martins, Marinho, and Martins}]{martins2019joint}
Pedro~Henrique Martins, Zita Marinho, and Andr{\'e}~FT Martins. 2019.
\newblock Joint learning of named entity recognition and entity linking.
\newblock In \emph{Proceedings of the 57th Annual Meeting of the Association
  for Computational Linguistics: Student Research Workshop}, pages 190--196.

\bibitem[{Mikolov et~al.(2013)Mikolov, Sutskever, Chen, Corrado, and
  Dean}]{mikolov2013distributed}
Tomas Mikolov, Ilya Sutskever, Kai Chen, Greg Corrado, and Jeffrey Dean. 2013.
\newblock Distributed representations of words and phrases and their
  compositionality.
\newblock In \emph{Proceedings of the 26th International Conference on Neural
  Information Processing Systems-Volume 2}, pages 3111--3119.

\bibitem[{Murty et~al.(2018)Murty, Verga, Vilnis, Radovanovic, and
  McCallum}]{murty2018hierarchical}
Shikhar Murty, Patrick Verga, Luke Vilnis, Irena Radovanovic, and Andrew
  McCallum. 2018.
\newblock Hierarchical losses and new resources for fine-grained entity typing
  and linking.
\newblock In \emph{Proceedings of the 56th Annual Meeting of the Association
  for Computational Linguistics (Volume 1: Long Papers)}, pages 97--109.

\bibitem[{Nguyen et~al.(2016)Nguyen, Theobald, and Weikum}]{nguyen2016j}
Dat~Ba Nguyen, Martin Theobald, and Gerhard Weikum. 2016.
\newblock J-nerd: joint named entity recognition and disambiguation with rich
  linguistic features.
\newblock \emph{Transactions of the Association for Computational Linguistics},
  4:215--229.

\bibitem[{Phan et~al.(2017)Phan, Sun, Tay, Han, and Li}]{phan2017neupl}
Minh~C Phan, Aixin Sun, Yi~Tay, Jialong Han, and Chenliang Li. 2017.
\newblock Neupl: Attention-based semantic matching and pair-linking for entity
  disambiguation.
\newblock In \emph{Proceedings of the 2017 ACM on Conference on Information and
  Knowledge Management}, pages 1667--1676.

\bibitem[{Poerner et~al.(2020)Poerner, Waltinger, and
  Sch{\"u}tze}]{poerner2020bert}
Nina Poerner, Ulli Waltinger, and Hinrich Sch{\"u}tze. 2020.
\newblock E-bert: Efficient-yet-effective entity embeddings for bert.
\newblock In \emph{Proceedings of the 2020 Conference on Empirical Methods in
  Natural Language Processing: Findings}, pages 803--818.

\bibitem[{Raiman and Raiman(2018)}]{raiman2018deeptype}
Jonathan Raiman and Olivier Raiman. 2018.
\newblock Deeptype: multilingual entity linking by neural type system
  evolution.
\newblock In \emph{Proceedings of the AAAI Conference on Artificial
  Intelligence}, volume~32.

\bibitem[{Robertson et~al.(2009)Robertson, Zaragoza
  et~al.}]{robertson2009probabilistic}
Stephen Robertson, Hugo Zaragoza, et~al. 2009.
\newblock The probabilistic relevance framework: Bm25 and beyond.
\newblock \emph{Foundations and Trends{\textregistered} in Information
  Retrieval}, 3(4):333--389.

\bibitem[{Sennrich et~al.(2016)Sennrich, Haddow, and
  Birch}]{sennrich2016neural}
Rico Sennrich, Barry Haddow, and Alexandra Birch. 2016.
\newblock Neural machine translation of rare words with subword units.
\newblock In \emph{Proceedings of the 54th Annual Meeting of the Association
  for Computational Linguistics (Volume 1: Long Papers)}, pages 1715--1725.

\bibitem[{Shi and Lin(2019)}]{shi2019simple}
Peng Shi and Jimmy Lin. 2019.
\newblock Simple bert models for relation extraction and semantic role
  labeling.
\newblock \emph{arXiv preprint arXiv:1904.05255}.

\bibitem[{Sil et~al.(2012)Sil, Cronin, Nie, Yang, Popescu, and
  Yates}]{sil2012linking}
Avirup Sil, Ernest Cronin, Penghai Nie, Yinfei Yang, Ana-Maria Popescu, and
  Alexander Yates. 2012.
\newblock Linking named entities to any database.
\newblock In \emph{Proceedings of the 2012 Joint Conference on Empirical
  Methods in Natural Language Processing and Computational Natural Language
  Learning}, pages 116--127.

\bibitem[{Wu et~al.(2020)Wu, Petroni, Josifoski, Riedel, and
  Zettlemoyer}]{wu2020scalable}
Ledell Wu, Fabio Petroni, Martin Josifoski, Sebastian Riedel, and Luke
  Zettlemoyer. 2020.
\newblock Scalable zero-shot entity linking with dense entity retrieval.
\newblock In \emph{Proceedings of the 2020 Conference on Empirical Methods in
  Natural Language Processing (EMNLP)}, pages 6397--6407.

\bibitem[{Yamada et~al.(2016)Yamada, Shindo, Takeda, and
  Takefuji}]{yamada2016joint}
Ikuya Yamada, Hiroyuki Shindo, Hideaki Takeda, and Yoshiyasu Takefuji. 2016.
\newblock Joint learning of the embedding of words and entities for named
  entity disambiguation.
\newblock In \emph{Proceedings of The 20th SIGNLL Conference on Computational
  Natural Language Learning}, pages 250--259.

\bibitem[{Yao et~al.(2020)Yao, Cao, and Pan}]{yao2020zero}
Zonghai Yao, Liangliang Cao, and Huapu Pan. 2020.
\newblock Zero-shot entity linking with efficient long range sequence modeling.
\newblock In \emph{Proceedings of the 2020 Conference on Empirical Methods in
  Natural Language Processing: Findings}, pages 2517--2522.

\end{thebibliography}

\appendix
\section{Examining model errors and predictions}
In tables \ref{example_1},\ref{example_2}, \ref{example_3} we show some example mentions and model predictions.

\begin{table*}[t!]
\begin{tabular}{lcc}
\hline
\textbf{Mention} & \multicolumn{2}{c}{
\parbox{35em}{... When he took it off , \textbf{a boy} working for Insector Haga swiped it and inserted the card \" Parasite Paracide \" into his Deck as he ran away with it . Jonouchi managed to catch the boy and met up with Anzu , Bakura and Sugoroku as he did so . ....}} \\
\hline
        & \parbox{10em}{Haga ' s helper ( manga )} & \parbox{22em}{Haga ' s helper ( manga ) Haga ' s helper was a boy employed by Insector Haga to help him cheat in a Duel against Katsuya Jonouchi . Biography . Haga promised the boy a rare card if he could sneak the card " Parasite Paracide " into Jonouchi ' s Deck . ... \newline} \\
        & \parbox{10em}{Queue cutter} & \parbox{22em}{Queue cutter The " queue cutter " is an elementary or junior high school kid , who appeared once in the manga . Biography . The boy cut past Yugi in a queue to a Capsule Monster Chess coin machine , outside Old Man Dentures store . ... \newline} \\
        & \parbox{10em}{Daichi} & \parbox{22em}{Daichi Daichi is one of the children in Crow ' s care in " Yu - Gi - Oh ! 5D ' s " . He has black hair in a bowl cut and dark blue eyes . ...} \\
\hline
\end{tabular}
\caption{\label{example_1} First example with the mention and the entity candidates. The correct entity candidate is the first from the candidate set.}
\end{table*}

\begin{table*}[t!]
\begin{tabular}{lcc}
\hline
\textbf{Mention} & \multicolumn{2}{c}{
\parbox{35em}{... He stole the Type 7 shuttlecraft " " , intending to join a freighter on Beltane IX and asked Captain Picard to tell \textbf{his father} he ' s sorry but had to do this . ...}} \\
\hline
        & \parbox{10em}{Jack Crusher} & \parbox{22em}{Jack Crusher Lieutenant Commander Jack R . Crusher was a Starfleet officer . Considered by Jean - Luc Picard to have been his best friend , he served under Picard ' s command on the . He was husband to Beverly Crusher and father to Wesley Crusher . ... \newline} \\
        & \parbox{10em}{Kurland} & \parbox{22em}{Kurland Kurland was a Human male aboard the in 2364 . He had one son , Jake Kurland . When Jake tried to leave the " Enterprise " - D with a shuttlecraft that year , he asked Captain Picard to tell his father he was sorry but he could not stay aboard the ship . Picard told him he should bring back the shuttle and tell this his father himself . ( ) ... \newline} \\
        & \parbox{10em}{Gabriel Lorca} & \parbox{22em}{Gabriel Lorca Captain Gabriel Lorca was a male Human Starfleet officer who lived during the mid - 23rd century . He served as the commanding officer on board at least one Federation starship , ...} \\
\hline
\end{tabular}
\caption{\label{example_2} Second example with the mention and the entity candidates. The correct entity candidate is the second from the candidate set.}
\end{table*}

\begin{table*}[t!]
\begin{tabular}{lcc}
\hline
\textbf{Mention} & \multicolumn{2}{c}{
\parbox{35em}{... Riker argues with him and is generally uncooperative . Remmick asks La Forge in engineering about \textbf{the incident} with Kosinski and the Traveler , and La Forge is forced to acknowledge that the captain lost control of the ship . ... }} \\
\hline
        & \parbox{10em}{USS Enterprise bridge holoprogram} & \parbox{22em}{USS Enterprise bridge holoprogram The USS " Enterprise " bridge holoprogram was a holodeck recreation of the bridge of the original . The program was accessed by Captain Montgomery Scott when he was on board the in 2369 after having been rescued from transporter stasis on the . ... \newline} \\
        & \parbox{10em}{Captain ' s log , USS Enterprise ( NCC - 1701 )} & \parbox{22em}{Captain ' s log , USS Enterprise ( NCC - 1701 ) The captain ' s log on the was the method used by the commanding officer to record the ship ' s events . These logs included Captain James T . Kirk ' s famous five - year mission . ( ) ... \newline} \\
        & \parbox{10em}{Traffic accident} & \parbox{22em}{Traffic accident A traffic accident was an incident involving vehicle s and their occupants in which the vehicle in question malfunctioned or crashed . Some accidents resulted in death . In 1930 , Edith Keeler died in a traffic accident after she crossed the street to find out why James T . Kirk had left her abruptly . ...} \\
\hline
\end{tabular}
\caption{\label{example_3} Third example with the mention and the entity candidates. The correct entity candidate is not part of the candidate set.}
\end{table*}

\end{document}